\documentclass[10pt,twocolumn,letterpaper]{article}

\usepackage{cvpr}
\usepackage{times}
\usepackage{epsfig}
\usepackage{graphicx}
\usepackage{amsmath}
\usepackage{amssymb}
\usepackage{pifont}% http://ctan.org/pkg/pifont
\newcommand{\cmark}{\ding{51}}%
\newcommand{\xmark}{\ding{55}}%
\usepackage[pagebackref=true,breaklinks=true,letterpaper=true,colorlinks,bookmarks=false]{hyperref}

\cvprfinalcopy % *** Uncomment this line for the final submission

\def\cvprPaperID{2250} % *** Enter the CVPR Paper ID here

\ifcvprfinal\fi
\begin{document}

%%%%%%%%% TITLE
\title{Regularizing Reasons for Outfit Evaluation with Gradient Penalty}

\author{Xingxing Zou\\
The Hong Kong Polytechnic Univeristy\\
{\tt\small aemika.zou@connect.polyu.hk}
% For a paper whose authors are all at the same institution,
% omit the following lines up until the closing ``}''.
% Additional authors and addresses can be added with ``\and'',
% just like the second author.
% To save space, use either the email address or home page, not both
\and
Zhizhong Li\\
The Chinese University of Hong Kong\\
{\tt\small lz015@ie.cuhk.edu.hk}
\and
Ke Bai\\
Duke University\\
{\tt\small ke.bai@duke.edu}
\and
Dahua Lin\\
The Chinese University of Hong Kong\\
{\tt\small dhlin@ie.cuhk.edu.hk}
\and
Waikeung Wong\\
The Hong Kong Polytechnic Univeristy\\
{\tt\small calvin.wong@polyu.edu.hk}
}

\maketitle
%\thispagestyle{empty}
%%%%%%%%% ABSTRACT
\begin{abstract}
    In this paper, we build an outfit evaluation system which provides feedbacks consisting of a judgment with a convincing explanation. The system is trained in a supervised manner which faithfully follows the domain knowledge in fashion. We create the EVALUATION3 dataset which is annotated with judgment, the decisive reason for the judgment, and all corresponding attributes (\eg print, silhouette, and material \etc.). In the training process, features of all attributes in an outfit are first extracted and then concatenated as the input for the intra-factor compatibility net. Then, the inter-factor compatibility net is used to compute the loss for judgment. We penalize the gradient of judgment loss of so that our Grad-CAM-like reason is regularized to be consistent with the labeled reason. In inference, according to the obtained information of judgment, reason, and attributes, a user-friendly explanation sentence is generated by the pre-defined templates. The experimental results show that the obtained network combines the advantages of high precision and good interpretation.
\end{abstract}

%%%%%%%%% BODY TEXT
\section{Introduction}\label{sec:intro}
Fashion compatibility evaluation is closely related to our daily life~(\eg Echo Look~\cite{echolook}),
and it has attracted increasing attention from researchers~\cite{cui2019dressing,Cucurull_2019_CVPR,wang2019outfit}.
Mainstream methods for fashion compatibility evaluation adopt metric learning:
% Outfits appear in the dataset are assumed to be positive while the randomly matched samples~\cite{han2017learning} are treated as negative~\cite{yu2018aesthetic}.
fashion items of the outfit are embedded into a common compatibility space,
% The items which appear in positive sample are closer otherwise have a greater distance.
where items that appear in the dataset are closer in representation and otherwise,
have a farther distance.
% In this way, only weak labels are required.
They assume that the occurrence rate of an outfit has direct relevance with its compatibility,
which effectively equates the concepts of common and uncommon to compatible and incompatible.
However, a discrepancy between \emph{being common} and \emph{being good} exists in fashion,.
A very common outfit is more likely to be normal rather than good.
How to provide professional evaluations that give convincing judgments of good, normal and bad,
is still open.

\begin{figure}[t]
    \begin{center}
        \includegraphics[width=1\linewidth]{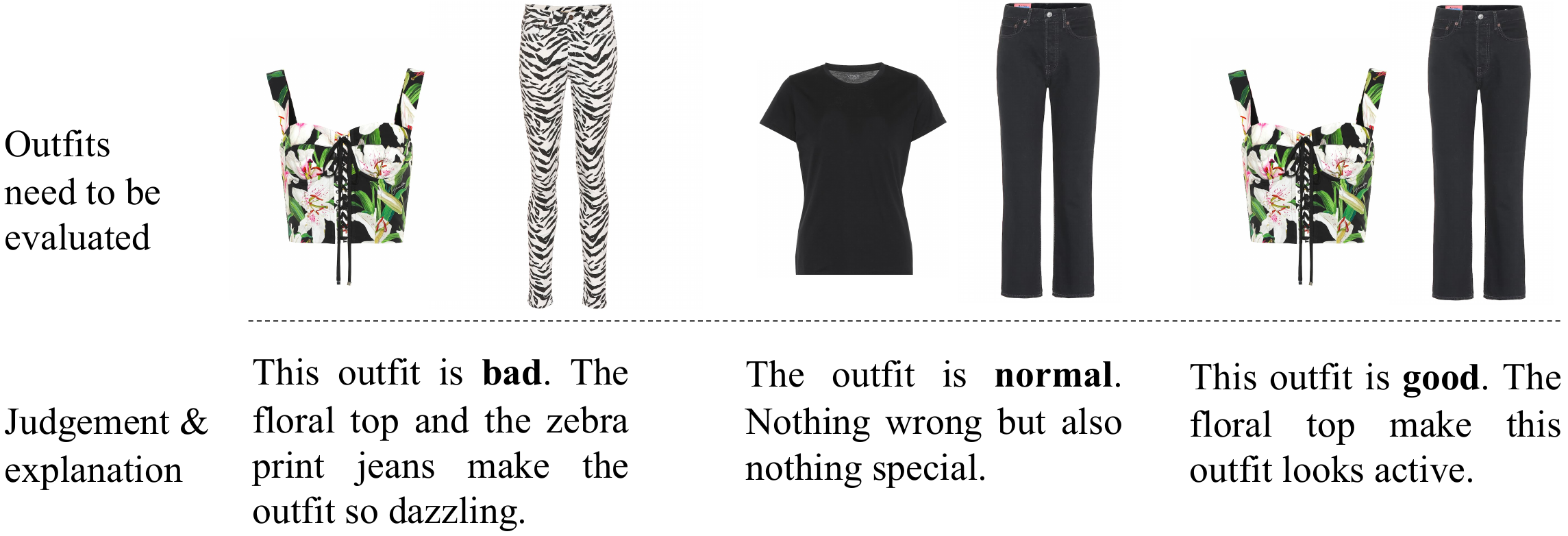}
    \end{center}
    \caption{\small
        Fashion compatibility evaluation with reason.
        Our evaluation system offers a clear judgment with convincing explanations.}\label{fig:work}
\end{figure}

A few efforts further focused on giving explanations for the output judgments.
    {}\cite{lin2018explainable,chen2018visually} took user reviews as training data to generate textual explanations.
    {}\cite{hou2019explainable} analyzed the outfit images and used heat maps as their explanation.
    {}\cite{tangseng2019toward} decomposed the item images into human-interpretable features,
and identified the most influential feature that contributes to the output.
However, the explanation generated by those methods is short on convincingness because of the following limitations:
1.~Not relate to specific and concrete reasons.
The textual explanation could be very vague,
\eg``This dress is so beautiful, I like it''.
2.~Lack of domain expertise.
It might simply recognize fashion attributes rather than analyze their relation,
\eg ``This orange T-shirt and black pants''.
3.~Not aligned with human experience.
Heatmaps may attend to image regions that are hard for the human to comprehend.

In this work, we build an outfit evaluation system that
faithfully respects the domain knowledge in fashion.
The judgment is summarized into three levels:
the outfit is good, bad, or normal.
For example, the first outfit in Figure~\ref{fig:work} looks bad because the unmatched print types between the top and bottom make its appearance too dazzling.
The mismatch in print is the logical reason to form the judgment.
% Similarly, the third outfit is good for the prints match with active feelings.
The \normal is a common situation when evaluating outfits.
For example the second outfit in Figure~\ref{fig:work},
an ordinary black T-shirt with black jeans, does not reach the bar of a good mix and match, but not bad either.
While it is hard to single out a concrete reason for being normal,
we can give an explanation for the judgment of good and bad.
In fashion, the major factors for evaluating outfits include color, print, material, \etc.~\cite{eckman1995aesthetic}.
The judgment is based on the overall visual expression of those interplaying factors.
An outfit is regarded as bad as long as one factor is not well-matched.
% In other words, not all factors in a bad outfit are incompatible.
% Meanwhile, only the attributes belong to the same factor could do the evaluate them match or not.
% For example, it is illogical to say white (color) and chiffon (material) match or not match.
If all factors arrive at visual harmony,
then the outfit at least can be put into normal.
Moreover, a good one must possess certain special design to make it stand out from a normal one.
In summary, normal is the intermediate level between good and bad;
%  and does not need explanation;
and for good and bad, we further expect a concrete reason for its deviation from normal.
% The reason is attributed to the aforementioned factors.
% And logically, the judgment of each outfit has a reason attributed to some factor.

In correspondence to the above discussion, we prepared a dataset that consists of outfit images annotated with the judgment and its decisive reason.
To realize the evaluation with reason task, we propose an explainable evaluation framework as follows.
Given an outfit composed of several fashion items, feature vectors for color, print, material, silhouette, and design details are first extracted.
For each of the factors, an independent net is used to produce an intra-factor compatibility feature.
Then all intra-factor compatibility features are concatenated and input into the inter-factor compatibility net, which outputs the judgment.
Reason for the judgment is traced back by computing gradients of the judgment \wrt the previous intra-factor compatibility features, in a way that resembles Grad-CAM~\cite{selvaraju2017grad}.
To increase reasonableness other than simply interpreting as-is, we enforce the traced reason to align with annotated reason by adding a regularization in the form of gradient penalty.
%  rather than simply interpret as-is.
% All in all, we not only let the network classify the judgment correctly in the forward pass but also make it trace back to the `correct' reason in the backward pass.
Based on the results obtained by the network, a user-friendly explanation is generated based on the pre-designed decision tree.

Our main contributions include:
(1) We formulate the fashion compatibility evaluation with reason task in a new framework, which respects the domain knowledge in fashion.
(2) We annotate an outfit evaluation dataset EVALUATION3 which has 18,108 pairs of outfit with judgments, reasons and attributes.
(3) We use gradient penalty to align the explanation of the network decisions to that of expert's.
Dataset and source code will be released shortly.

\section{Related Work}\label{sec:related}
\paragraph{Visual compatibility learning.}
Simo-Serra \etal~\cite{simo2015neuroaesthetics} learned the fashionability using Conditional Random Field.
Oramas \etal~\cite{oramas2016modeling} adopted mid-level attributes to measure the compatibility.
Li \etal~\cite{li2017mining} used annotated quality scores to supervise the grading of outfits.
Han \etal~\cite{han2017learning} used Bi-LSTM,
and Visileva \etal~\cite{vasileva2018learning} improved on it by considering type information.
Recent works built upon the clothing embedding~\cite{hsiao2018creating,wang2019outfit},
which can be learned through autoencoder or by the Bayesian Personalized Ranking~(BPR) loss~\cite{song2017neurostylist},
the hinge loss~\cite{kang2018complete},
the triple loss~\cite{chen2018dress},
and the binary cross-entropy loss~\cite{shih2018compatibility}.
Another line of work models the clothing style~\cite{mcauley2015image,veit2015learning,hsiao2017learning,al2017fashion},
which expresses the compatibility implicitly.
In a word, mainstream approaches mainly adopt the relative embeddings to compute evaluation scores.
The benefit is not needing explicitly graded data for they assume all outfits in the dataset are positive samples.
However, it is unreasonable to equate the observation of \emph{high occurrence rate} with the concept of \emph{good}.
Meanwhile, a relative score is less friendly to users and the compatibility space is hard to interpret.
Our work differs from above in several aspects.
Firstly, we adopt the fashion-matching principles~\cite{eckman1995aesthetic} as the standard for evaluating whether an outfit is good or bad.
Secondly, we propose absolute ratings with three levels as judgments in the evaluation scene.
Thirdly, we are interested in the explainability of the given judgment.

\paragraph{Explainable outfit recommendation.}
Yang \etal~\cite{yang2019interpretable} introduced an attribute-based interpretable compatibility method.
They mixed all attributes and found the informative attribute crosses statistically,
whereas we find the dominant factor by mimicking the analyzing process of fashion experts.
Feng \etal~\cite{feng2018interpretable} proposed a partitioned embedding network,
where color, shape and texture are defined as the main factors,
and the score of each factor is used as the explanation.
Lin \etal~\cite{lin2018explainable}
presented a recommendation system and used the generated comments as explanations.
Xu \etal~\cite{chen2018visually} adopted user reviews into an attention-based architecture to enhance the performance of recommendation and the interpretability.
These methods provide reasons using the heatmap, comments or reviews,
in contrast, our explanation is inferred based on the trained network.
Recently, Tangseng \etal~\cite{tangseng2019toward} defined an influence score of factors using gradients,
which has a similar idea to this work.
The major difference is that,
they analyzed the trained model \emph{post-hoc},
while we explicitly supervise the explanation of the model to be aligned with expert interpretations.

\paragraph{Explaining DNN classifiers.}
Works on explaining neural networks greatly improve the interpretability of the black-box deep models.
Springenberg \etal~\cite{springenberg2014striving} visualized CNN predictions by highlighting contributing pixels.
Zhou \etal~\cite{zhou2016learning} proposed the Class Activation Mapping (CAM) for visualization.
Selvaraju \etal~\cite{selvaraju2017grad} presented Grad-CAM which adopted the gradients of any target concept to enhance localization.
Chattopadhyay \etal~\cite{chattopadhay2018grad} further improved the performance with Grad-CAM++.
Regularizations can also be added~\cite{al2017contextual,plumb2019regularizing,melis2018towards} to improve the explanation quality.
However,
% as will be demonstrated later,
directly using the heatmap produced by these methods as the reason for fashion judgments is insufficient,
% It does not work
because the reason for the judgment is not a local region of an item image but the global factors such as color, print, and silhouette.
To overcome this limitation,
we trace the reason for the judgment back to human-interpretable factors,
rather than pixel-wise heatmaps.
Traditional methods like decision tree also own limited explainablility~\cite{kim2018introduction}.
They are not suitable for our task because
1.\ For any outfit, tree based methods give a decision path from root to leaf,
which is hard to locate a single, dominate factor that contributes to the decision.
2.\ In the tree construction, the space is split based on heuristics such as information gain which is not directly relevant.

\section{Dataset Construction}\label{sec:dataset}
Based on the perspective of fashion,
we grade an outfit into three progressive levels: \bad, \normal, and \good.
% the three levels are clearly defined.
The \bad level is defined as the outfit having something wrong,
\eg a wrong color matching or a dazzling print.
If the outfit does not make any mistake that breaks the visual balance,
it comes up to the \normal level.
Further, if the outfit has a special design,
\eg attractive color matching, special print, or good cutting,
it reaches the \good level.
Note the logical connection among the three levels.
If the outfit has something wrong, \ie some factors not well-matched,
it will be regarded as \bad, whether it has some special design or not.

Our evaluation system needs three-level judgment with the corresponding reason.
Detailed attribute annotations are also needed to train factor feature extractors.
However, existing public fashion datasets do not have labeled judgments and their decisive reasons.
Thus we introduce a new dataset, namely \evaluationthree.
The image source is a subset of the \emph{Polyvore dataset}~\cite{han2017learning}.
The dataset contains four sets including train set, validation set, test set, and test-random set.
All labels are manually annotated by 9 independent annotators (all major in fashion).
Voting mechanism was adopted to get the final label.
Furthermore, to mitigate the influence of cultural backgrounds,
the 9 annotators (4 male and 5 female) were selected from different fashion regions including Asia-Pacific, Europe, North America, \etc.

\begin{figure}[t]
    \begin{center}
        \includegraphics[width=1.0\linewidth]{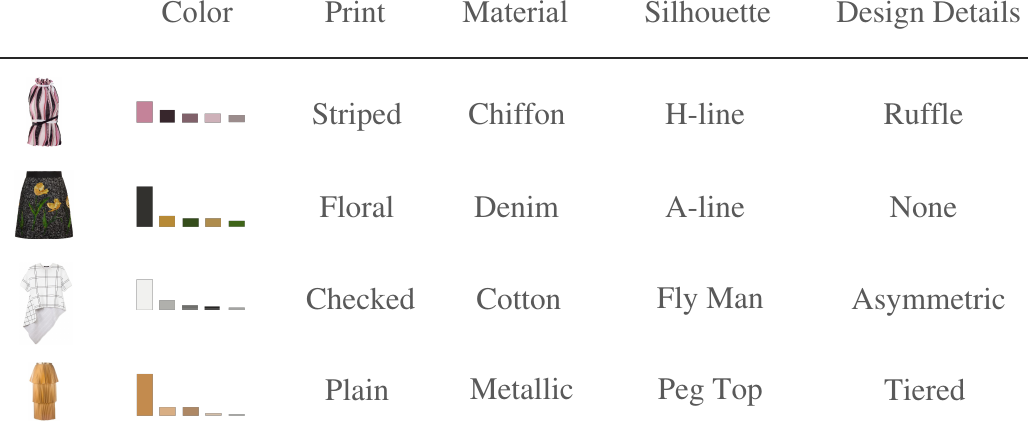}
    \end{center}
    \caption{\small
        Samples of labeled attributes in the \evaluationthree, which cover material, silhouette, and design details.
        We also show the five main colors, and their histograms.
    }\label{fig:print14}
\end{figure}

Each outfit contains a top image and a bottom image.
We annotated 18,108 outfits in total, including 2,861 (15.8\%) in good, 13,587 (75.0\%) in normal,
and 1,660 (9.2\%) in bad.
Among them, 12,608 outfits (15.7\% in good, 75.4\% in normal, and 8.8\% in bad) belong to train set,
2,500 outfits (17.0\% in good, 74.4\% in normal, and 8.6\% in bad) are put into validation set,
and the rest 3,000 outfits (15.6\% in good, 73.8\% in normal, and 10.6\% in bad) are used for testing.
In particular, we added a \emph{test-random} set that randomly pairs tops and bottoms in the test set.
It contains 3,000 outfits with 22.8\% in good, 61.2\% in normal, and 15.4\% in bad.
This set provides a different data distribution because outfits in the original dataset are all paired by Internet users.

As our outfit evaluation is based on features of the design factors,
the accuracy of attribute recognition directly affects the evaluation result.
To reduce domain gap,
we labeled the corresponding attributes on our dataset instead of adopting different fashion attribute datasets.
We summarized 14 print types,
including abstract, allover, animal print, \etc.
Besides, there are 5, 8 and 10 attributes in silhouettes (A-line, H-line, Peg-top, \etc.),
design details (tiered, wrap, ruffle \etc.), and materials (knit, lace, leather, \etc.), respectively.
Figure~\ref{fig:print14} shows some examples of the dataset.

%%%%%%%%%%%%%%%%%%%%%%%%%%%%%%%%%%%%%%%%%%%%%%%%%%%%%%%%%%%%%%%%%
% APPROACH
%%%%%%%%%%%%%%%%%%%%%%%%%%%%%%%%%%%%%%%%%%%%%%%%%%%%%%%%%%%%%%%%%
\section{Approach}\label{sec:approach}

Given an outfit that comprises of a top and a bottom,
we first classify it into one of the three levels: \good, \normal, and \bad.
Except for \normal, a reason will then be identified among three choices: \rcolor, \rprint, and \rattribute.
Here \rattribute is a collective name for \emph{material}, \emph{silhouette}, and \emph{design details}.
The obtained results, including judgment, reason, and attributes, together generate the final explanation based on a pre-designed decision tree.
For convenience,
we denote the judgment set by \(\cJ\) = \{\emph{good}, \emph{normal}, \emph{bad}\},
the reason set by \(\cR\) = \{\rcolor, \rprint, \emph{design}\}.
% and the factor set by \(\cF\) = \{\rcolor, \rprint, \emph{material}, \emph{silhouette}, \emph{design details}\}.

\subsection{Architecture}
\begin{figure}[t]
    \begin{center}
        \includegraphics[width=1\linewidth]{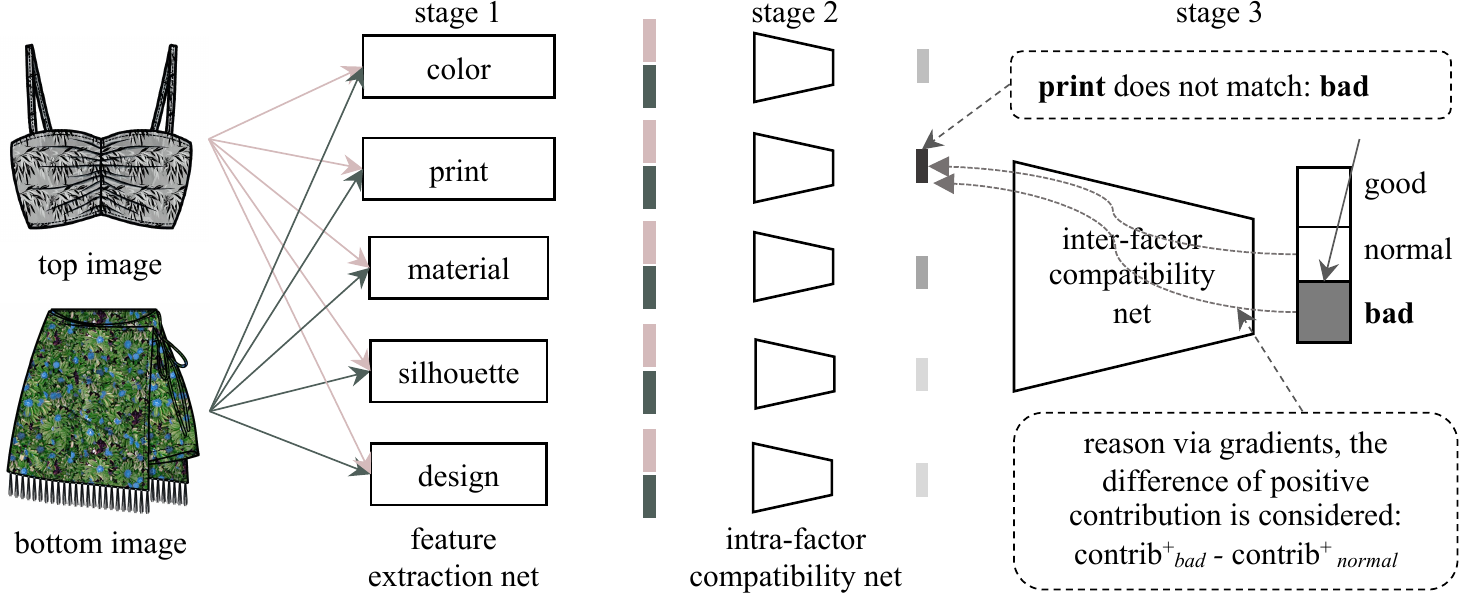}
    \end{center}
    \caption{\small
        The pipeline of outfit evaluation network.
        The features of each contributing factor are extracted by the feature extraction net.
        Then, the judgment and reasons are trained in the same network.}\label{fig:pipeline}
\end{figure}

% The whole framework can be divided into two parts: 1. Outfit evaluation with the neural network;
% 2. Explanation generation with pre-designed decision tree;
% For the first part,
The pipeline of computing judgment and reason shown in Figure~\ref{fig:pipeline}.
The judgment is obtained in three stages, namely
(1) factor feature representation,
(2) intra-factor compatibility and
(3) inter-factor compatibility.
For each pair of outfit images,
we attain the feature representation via the respective function or convolutional neural networks.
Then the features of each factor first pass through the intra-factor compatible net,
and produce the intra-factor compatible feature which mixes the up and bottom information.
At last, we use these intra-factor compatible features for the final judgment.
The contribution of factors to the decision is traced back via gradients to the intra-factor compatible feature,
and we can calculate the prominent reason for the judgment.
To generate user-friendly explanations,
we added a post-processing step using a sentence template, which is completed by the elements from a pre-designed decision tree.
Note that our evaluation framework can be extended to process the varied number of fashion items simply by replacing the CNNs in stage 2 to LSTMs~\cite{han2017learning}.
In the following, we will discuss the network design,
the process of tracing back reasons and the way of explanation generation in detail.

\paragraph{Fashion feature extraction.}
For each input image,
we have five feature vectors based on the five fashion compatibility evaluation factors, \ie color, print, material, silhouette, and design details.
Features of the print, material, silhouette and design details are extracted by deep models.
We fine-tune the ImageNet pre-trained ResNet-18~\cite{he2016deep} on the \evaluationthree.
The last feature map (\(512\)-dimensional) is used as the representation of these fashion factors.
% Specifically, t
Take the print feature net for example.
We
1.~change the output neurons of the last fc layer of ResNet from 1,000 to 14;
2.~initialize parameters using pre-trained ImageNet models;
3.~train the last fc layer on the dataset for 10 epochs with other parameters fixed;
4.~jointly train all parameters for 30 epochs.
For the color feature, we compute the color histogram,
The \emph{Fashion Color system (FOCO)}~\cite{zou2019foco} is used to set up bins of the histogram.
FOCO divides the \emph{H}, \emph{S}, \emph{B} channels into 15, 8 and 6 levels, respectively.
For example, PANTONE~\cite{pantone} Candy Apple Red is represented as (1, 8, 4) in FOCO\@.
We concatenate the five major colors with their ratio and obtain a \(25\)-dimensional color feature.
% All in all,
% the color feature vector is \(25\)-dimensional,
% and the rest 4 are \(512\)-dimensional vectors.

\paragraph{Intra-factor compatibility network.}
% After feature extraction, f
For each outfit and each factor, we obtain two feature vectors corresponding to the top and the bottom.
Then, the two vectors are fused by an intra-factor compatibility network,
which is a three-layer fully-connected network.
The output feature represents the interaction between the top and the bottom on a specific factor.
In the end, all five output features are concatenated as a single intra-factor compatibility feature vector and sent to the next stage.
This intra-factor compatibility feature is also the place where we trace the reason back to.

\paragraph{Inter-factor compatibility network.}
This network (also has three fully-connected layers) captures how all factors are inter-related to the final decision.
The input is the five intra-factor compatibility features,
and the output is the probability of being judged as \good, \normal or \bad.
% It is also a three-layer fully connected network.
% The output feature of inter-factor compatibility network is
The cross-entropy loss \(L_{\text{judgment}}\) for judgments is computed therefrom.

\subsection{Reason for Judgments}
Grad-CAM~\cite{selvaraju2017grad} is widely used for inspecting the spatial contributing regions of inputs.
%  and analyzing factor contributions.
Conditioned on a class label,
it finds the dominant neurons contributing to the decision toward this label by tracking back the gradient of the label logit (\ie logit before softmax) \wrt feature maps.
For any class label \(c\),
the gradient of the logit \(y_c\) with respect to the \(k\)-th feature map \(A^k\) of a convolutional layer, is computed.
Then these gradients are global-average-pooled to obtain the neuron importance weights \(\alpha_k^c\),
\begin{equation}\label{eq:weight}
    \alpha_k^c := \frac{1}{Z}\sum_{p}\sum_{q}\frac{\partial y_c}{\partial A_{pq}^k},
\end{equation}
where \(p, q\) iterates over the spatial dimensions and \(Z\) is the number of pixels in the feature map.
The product of the neuron importance weights and the forward activation map is obtained as the heatmap \(H^{\text{Grad-CAM}}_c\),
\begin{equation}\label{eq:loss0}
    H^{\text{Grad-CAM}}_c := \relu\left(\sum_{k}\alpha_k^c A^k\right).
\end{equation}

Different from Grad-CAM,
we care about the ``heatmap'' over human-interpretable factors rather than images pixels.
Denote the logit for the judgment \(j\in\{\text{good},\text{normal},\text{bad}\}\) by \(y_j\).
Let the intra-factor compatibility feature be \(\vx\) with elements \(x_i\).
We define the \emph{contribution} of each element \(x_i\) for the decision of judgment \(j\) as \(\contrib_j\),
\begin{equation}\label{eq:contrib}
    \contrib_j(x_i) := \frac{\partial y_j}{\partial x_i} \odot \relu(x_i).
\end{equation}
%
% where \((x)_+\) is a shorthand for \(\max{(x, 0)}\).
% Neurons in the feature map \(\vx\) has semantic meanings.
As shown in Figure~\ref{fig:pipeline},
the intra-factor compatible feature is divided into five distinct parts,
each represents the compatible situations of one factor, such as color or print.
Suppose the index set of neurons for factor \(r\in\{\text{color}, \text{print}, \text{design}\}\)
is \(I_r\),
then we define the contribution of the factor as the average of the contribution of the constituent elements.
and the \emph{positive contribution} of this factor for the judgment \(\cJ\) as

\begin{equation}\label{eq:contrib_plus}
    \contrib_j^+ (r) := \frac{1}{|I_r|} \sum_{i \in I_r} \relu\bigg(\frac{\partial y_j}{\partial x_i}\bigg) \odot \relu(x_i).
\end{equation}

The positive contribution in Equation~\eqref{eq:contrib_plus} is slightly different from the form of Grad-CAM,
in which the \(\relu\) is put inside the summation.
Unlike Grad-CAM, whose goal is to analyze the network after training,
we actively regularize the network to be explainable during the training.
Formulation of Equation~\eqref{eq:contrib} and~\eqref{eq:contrib_plus} is based on the following contribution assumption;
we also introduce the relativeness assumption.

(1) \emph{Contribution assumption:
    the contribution of a neuron \(x_i\) for the decision consists of two parts:
    the saliency and the sensitivity.}
Firstly the value should be present,
and the impact is greater if the value is larger.
If it is negative, then it will be filtered out immediately by the subsequent \(\relu\) activation.
Secondly, the decision should be sensitive to the change of the neuron,
which is measured by the derivatives~\cite{simonyan2013deep}.
If the value is large but the derivative is zero,
it means that the decision is irrelevant to the presence of the neuron.
% A negative derivative would imply a negative contribution.
The product of saliency and sensitivity has been used previously,
such as in the Grad-CAM, the Item Feature Influence Value (IFIV)~\cite{tangseng2019toward},
and the \emph{relevance score} in~\cite{montavon2018methods} which is derived from the perspective of Taylor decomposition.

(2) \emph{Relativeness assumption:
    the difference between two contributions has semantic meaning.}
This motivates from the fashion domain knowledge:
both the \good and \bad levels are defined relative to the \normal level.
What people care the most for the \good is its highlight comparing to a normal one,
rather than a full list of the contributing factors.
Good color matching,
nice print combination, \etc., all may contribute to its being good but not necessarily stand out.
The same argument applies to the \bad level.
% Each factor contributes to both good and normal.
We aim to find the factor which contributes the most to \good compared with \normal.
Applying the relativeness assumption specifically to the evaluation task at hand,
the mathematical form of the reason for the \good  judgment can be formulated as
\begin{equation}\label{eq:reason_good}
    r_{\text{good}} := \argmax_{r\in \cR} \big(\contrib^+_{\text{good}}(r) - \contrib^+_{\text{normal}}(r)\big).
\end{equation}
We use the good minus normal to estimate to which extent will each factor lead to good rather than normal,
and use the \(\argmax\) operator to extract the dominant factor.
Similarly, \bad is bad because it is worse than normal.
The reason \(r_{\text{bad}}\) is
\begin{equation}\label{eq:reason_bad}
    r_{\text{bad}} := \argmax_{r\in \cR} \big(\contrib^+_{\text{bad}}(r) - \contrib^+_{\text{normal}}(r)\big).
\end{equation}
For \normal cases,
there is no explicit reason available because all factors are neither outstanding nor terrible.

\subsection{Supervise Reason with Gradient Penalty}\label{sec:supervise}

We want the prominent reason output by the network to be aligned with pre-labeled data.
This is achieved by training the network with specially designed regularizations.
%
% The regularizer is to make the reason prediction to conform with the ground-truth reason.
We propose three forms of regularizations and compare their performances in the experiment section.
Let \(F=(F_{color}, F_{print}, F_{design})\) be the traced reason vector for the ground-truth class, where
\begin{equation}\label{eq:f}
    F_r := \sum_{j\in\mathcal{J}} \doubleone_{j^{gt}}(j) \cdot \contrib^+_{j}(r) - \contrib^+_{\text{normal}}(r).
\end{equation}
Here \(r\in\cR\), \(\doubleone_{j^{gt}}\) is an indicator function for ground-truth judgment \(j^{gt}\in\cJ\).
When judgment \(j\) is the same as ground-truth, \(\doubleone_{j^{gt}}(j)=1\);
otherwise, \(\doubleone_{j^{gt}}(j)=0\).
\eg \(\doubleone_{\text{good}}(j)=1\) if \(j\) represents good.
The three components of \(F\) are the reason strength of factor \rcolor, \rprint and \rattribute, respectively.
Denote the ground-truth reason as \(r^{gt}\),
then the three regularizations can be expressed as the following.
The \emph{reason} shown in the formulation below refers to the label of reason, \ie \rcolor, \rprint and \rattribute, respectively.

\begin{enumerate}
    \item Cross-entropy (CE) regularizer,
          \begin{equation}
              L_{\text{reason}}^{\text{CE}} = -\log\left(\frac{\exp(F_{r^{gt}})}{\sum_{r\in\cR} \exp(F_r)}\right).
          \end{equation}
    \item Linear regularizer. If the prediction is wrong, then linearly pull up the ground-truth reason strength and push down the wrong one,
          \begin{equation}
              L_{\text{reason}}^{\text{linear}} = \max_{r\in\cR}\big(F_r\big) - F_{r^{gt}}.
          \end{equation}
    \item Square regularizer, %The squared version of linear regularization.
          \begin{equation}
              L_{\text{reason}}^{\text{square}} = \left(\max_{r\in\cR}\big(F_r\big) - F_{r^{gt}}\right)^2.
          \end{equation}
\end{enumerate}

The total loss \(L\) is the combination of the judgment loss and the reason loss, \ie one of the above regularizers,
\begin{equation}\label{eq:loss}
    L = L_{\text{judgment}} + \alpha L_{\text{reason}},
\end{equation}
where \(\alpha\) is a hyper-parameter that controls the effect of reason regularization.
The \(L_{\text{judgment}}\) and \(L_{\text{reason}}\) is the loss of the judgment and reason, respectively.
Since the gradient appears in the definition of contribution and reason (Equations~\eqref{eq:contrib_plus} and~\eqref{eq:f}),
the loss term \(L_{\text{reason}}\) penalizes the gradient.
The gradient penalty directly affects the network parameters after the compatible feature layer.
In the definition, feature map \(x\) also presents.
This means that the parameters of the intra-factor compatible net,
those before the compatible feature layer, are also regularized.

\subsection{Explanation Generation}
\begin{figure}[t]
    \begin{center}
        \includegraphics[width=1\linewidth]{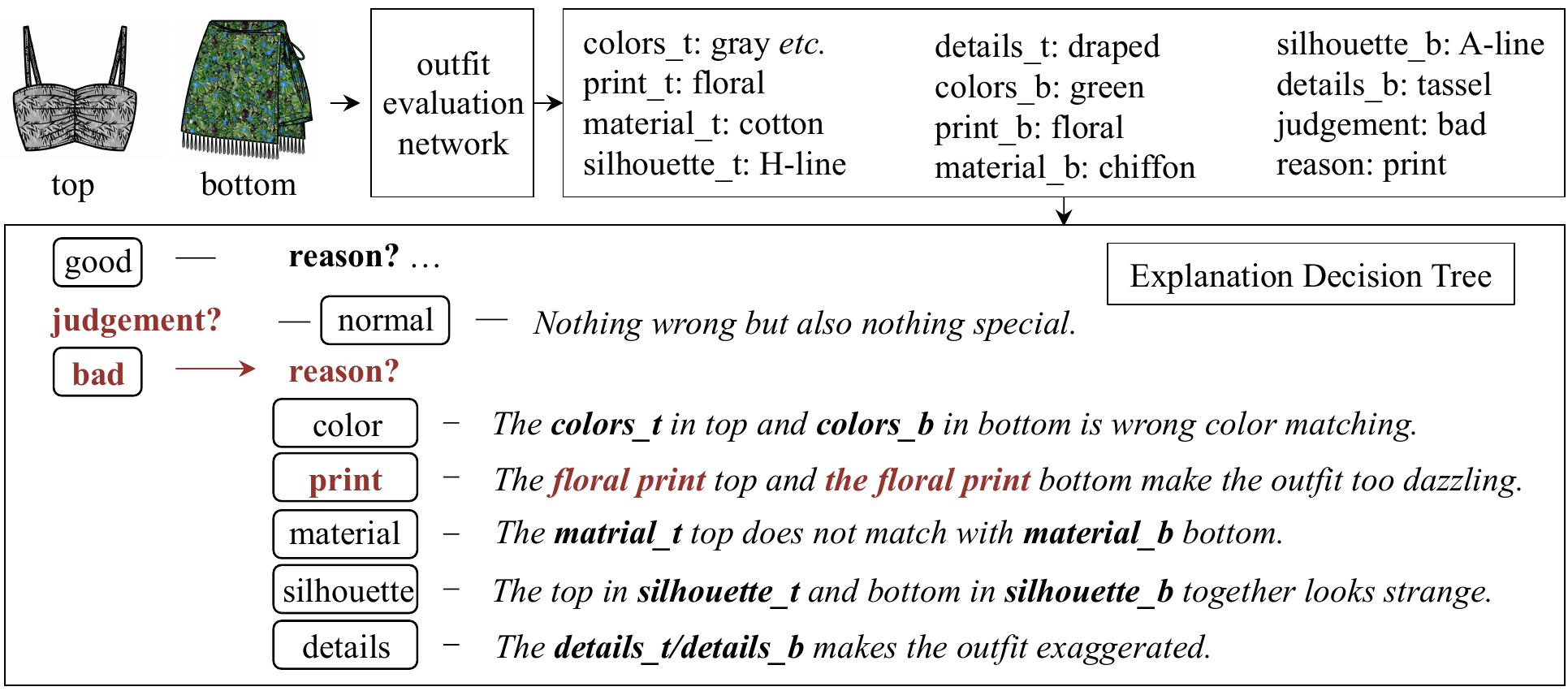}
    \end{center}
    \caption{\small
        The pipeline of explanation generation with the obtained judgment,
        reason and attributes of an outfit.}\label{fig:tree}
\end{figure}

To obtain explanations that are friendly to users,
we used an explanation template in the form of the decision tree (manually designed, not learnable, see Figure~\ref{fig:tree}),
which is built with the assist of fashion experts.
This part can also be implemented using different approaches such as text generation techniques.
% which is not the technical focus in this work.
The template is sufficient for our demonstration purpose.
After analyzing an outfit as described earlier,
we obtain the results including judgment, its decisive reason, and all corresponding attributes of the outfit.
Then,
the explanation is generated accordingly following the decision tree.
For example, as shown in Figure~\ref{fig:tree},
the judgment is \bad while the predicted reason is print.
From the prescribed sentence template:
``The \emph{print\_t} top and the \emph{print\_b} bottom make the outfit too dazzling'',
we can obtain the explanation:
\emph{``This outfit is bad.
    The floral print top and the floral bottom make the outfit too dazzling.''}
% Noted that our explanation is not only attribute-based for each outfit but also concluded from its decision factor.
% The floral print top and the floral bottom are incompatible (\textit{conclusion}) because of the two prints make the whole outfit too dazzling (\textit{reason}).
We found that it is unnecessary to mentioned attributes other than the main factor,
\eg the short-length of the skirt, because they have little effect on whether the outfit is compatible.

\begin{table*}[t]
	\begin{center}
		\begin{tabular}{l|c|c|c|c|c|c|c|c}
			\hline
			                 & judge & reason  & partition & explain & test-acc-j & test-acc-r & test-r-acc-j &test-r-acc-r \\
         \hline\hline
         Multi-CLS-Part                 & \cmark                        & \cmark               & \cmark               & \xmark                            & \(74.4\pm 0.5\)                          & \(74.8\pm 3.1 \)         & \(69.4 \pm 0.4\)         & \(75.3\pm4.6 \)          \\
         % Multi-CLS-Img                  & \cmark                        & \cmark               & \xmark               & \xmark                            & \(73.3\pm 1.5 \)                 & \(76.4\pm 1.4\)          & \boldmath\(71.7\pm 0.7\) & \(75.0\pm 0.2\)          \\
         % Partition-BPR                  &                               &                      & \cmark               & \cmark                            &                                  &        &            &        \\
         IFIV~\cite{tangseng2019toward} & \cmark                        & \xmark               & \cmark               & \cmark                            & \(73.3\pm 0.7\)                          & \(35.9\pm 4.5\)          & \(69.7\pm 0.5\)          & \(37.2\pm 5.7\)          \\
         Reason-NoReg                   & \cmark                        & \xmark               & \cmark               & \cmark                            & \(73.3\pm 0.7\)                          & \(38.9\pm 3.2\)          & \(69.7\pm 0.5\)          & \(40.1\pm 4.1\)          \\
         Ours linear                    & \cmark                        & \cmark               & \cmark               & \cmark                            & \(72.8\pm 1.3\)                          & \(68.3\pm 2.4\)          & \(69.1\pm 0.5\)          & \(69.9\pm 4.2\)          \\
         Ours square                    & \cmark                        & \cmark               & \cmark               & \cmark                            & \(72.1\pm 1.6\)                          & \(73.8\pm 1.6\)          & \(68.8\pm 0.3\)          & \(74.9\pm 0.9\)          \\
         Ours cross-entropy             & \cmark                        & \cmark               & \cmark               & \cmark                            & \boldmath\(74.8\pm 1.7\)                 & \boldmath\(76.7\pm 3.6\) & \boldmath\(70.0\pm 0.2\) & \boldmath\(76.8\pm 0.8\) \\
			\hline
		\end{tabular}
	\end{center}
	\caption{\small
        Comparing our method with related methods.
        All experiments were repeated 5 times.
        The first line is multi-task (judgment and reason) classification models,
        where \emph{Multi-CLS-Part} uses the same fashion-related features as our model.
        Both \emph{IFIV} and \emph{Reason-NoReg} did not take the supervision of reason.
        For \emph{IFIV},
        we did not implement temperature-scaling,
        since only the factor with maximum contribution was considered.
        The rest are our method with three different regularizations.
    }\label{tab:compare}
\end{table*}

\section{Experiment}\label{sec:exp}

\begin{figure*}[t]
    \begin{center}
        \includegraphics[width=1\linewidth]{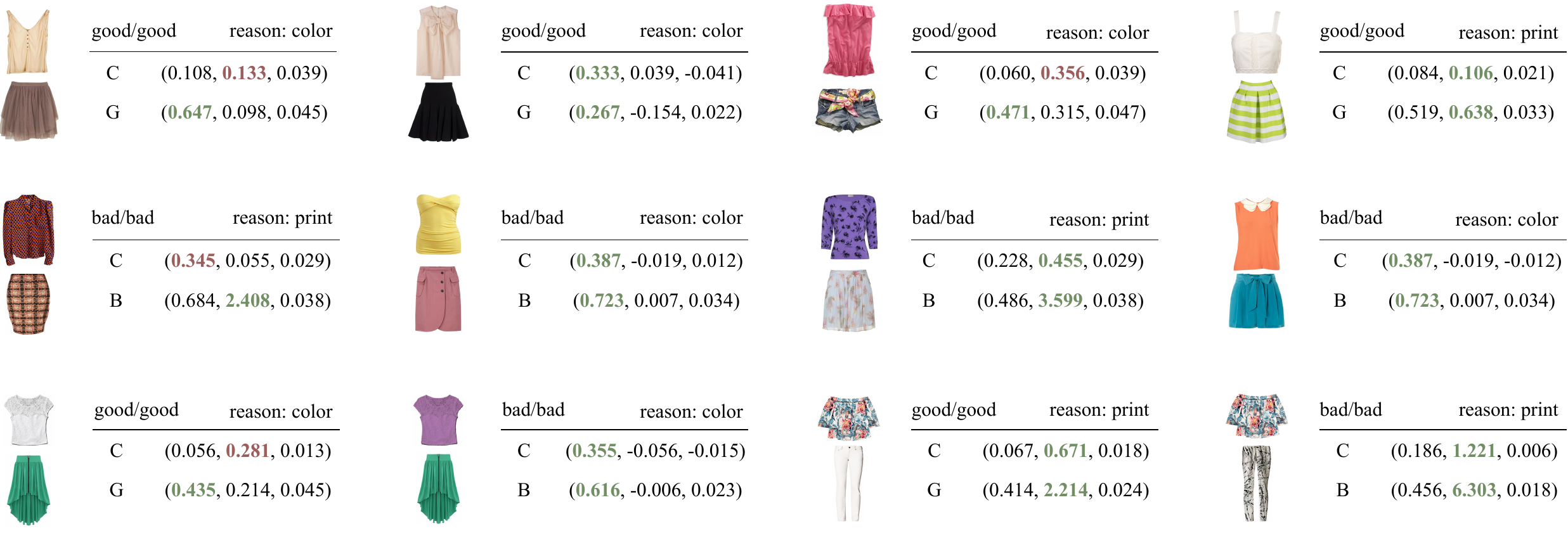}
    \end{center}
    \caption{\small
        The reason and contribution factor analysis on some outfits via a model with cross-entropy reason-regularization.
        At the top of each mini-table, \emph{good/good} means ground-truth/predicted judgment,
        and the ground-truth reason is also shown.
        In each mini-table,
        the three columns represent the component of factors \rcolor, \rprint and \rattribute, respectively.
        Rows with \emph{C} are the contribution (see Equation~\eqref{eq:contrib}) of predicted judgment,
        \emph{G} are the reason for \good~(see Equation~\eqref{eq:reason_good}) and \emph{B} are reason for \bad (see Equation~\eqref{eq:reason_bad}).
    }\label{fig:results}
\end{figure*}

\begin{figure}[t]
    \begin{center}
        \includegraphics[width=1\linewidth]{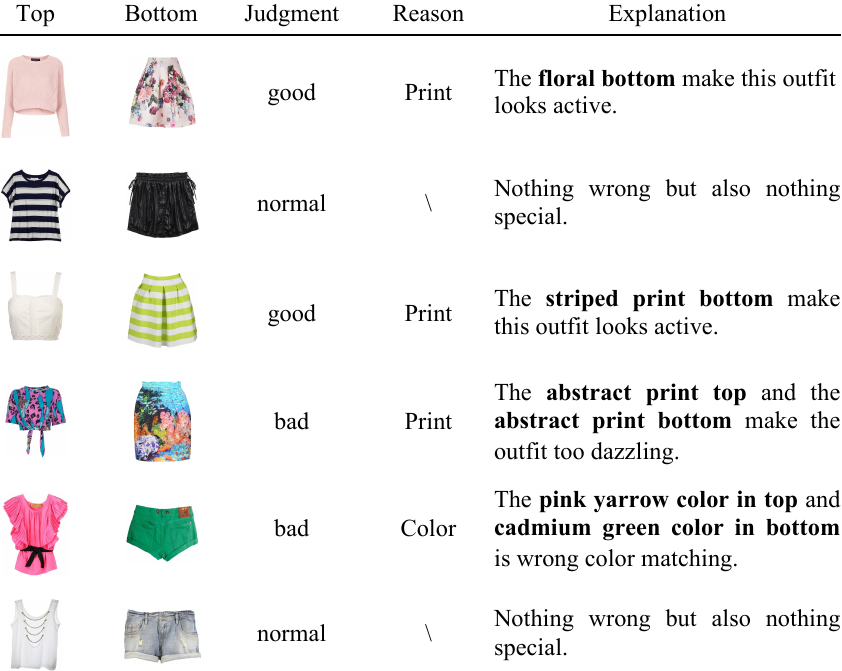}
    \end{center}
    \caption{\small
        The samples of our fashion compatibility evaluation including judgment, reason and explanation.}\label{fig:visual}
\end{figure}

In this section,
we first conduct experiments on \evaluationthree to show the advantages of the proposed evaluation system quantitatively and qualitatively.
The quantitative analysis focuses on both the judgment accuracy and the reason accuracy.
We show the superiority of our approach on the test set and the test-random set.
Recall that the test-random set is the random composition of fashion items in the test set,
which could be regarded as an indicator of the model's generalization ability.
Regarding qualitative analysis,
we visualize the detailed contributions in varied situations, \eg judgment:
\bad, reason: print, to verify the advantage of our method with gradient penalty.
Then, we provide ablation study on the choice of contribution formulations and regularizers.

\subsection{Compatibility Evaluation with Reasons}
\paragraph{Metric.}
We measure both the \emph{judgment accuracy} and the \emph{reason accuracy}.
The \emph{judgment accuracy} is computed as the number of correct predicted judgment (compared with the labeled ground-truth of judgment) divided by the number of total predicted samples.
It is worthwhile to note that, for an outfit, if the judgment is wrong, then the reason for this judgment may not make sense.
Thus, the premise of the correctly predicted reason is that the judgment is correct.
Under the condition of the right predicted judgment, we count the number of correctly predicted reason.
Then divided it by the number of total predicted outfits.
Then divide it by the number of correct predicted judgments.

\paragraph{Training details.}
We train the pipeline (Figure~\ref{fig:pipeline}) in two steps.
In the first step,
we train four CNNs to learn the representations of the print, the material, the silhouette and the design details separately.
The CNN structure is ResNet18.
Each factor has its corresponding labels, with examples shown in Figure~\ref{fig:print14}.
All the networks are trained using RMSProp with initial learning rate \(0.0001\) and weight decay \(0.00005\),
the learning rate is divided by 10 every 30 epochs.
Parameters of the pre-trained model are fixed for the first 20 epochs.
We use the feature of the second last fc-layer from the pre-trained model as the input to the second stage.
The color feature is directly extracted by the histogram.
In the second step,
we train the intra-factor compatibility networks,
the inter-factor compatibility network,
and the reason regularization (Modern DL frameworks such as PyTorch have built-in support for computing the gradient of gradients) jointly.
In particular, we use SGD with initial learning rate \(0.01\),
and weight decay \(0.0005\) for \(70\) epochs and the learning rate is also divided by \(10\) every 30 epochs.
Since the dataset is unbalanced (the ratio of \normal in the train set reaches 75.4\%),
we adjust the sampling rate for each class (\good, \normal, and \bad) to obtain the balanced sampled data.

\paragraph{Quantitative analysis.}
Table~\ref{tab:compare} shows the comparison on the accuracy with related methods.
\emph{IFIV}~\cite[Equation 10]{tangseng2019toward} computes the item feature influence value,
which has the same maximum factor as our \(\contrib_j\).
We also test the performance without regularization (\ie, set \(\alpha=0\)) as shown in \emph{Reason-NoReg}.
Meanwhile, we evaluate the performance with three regularizations, linear, square, and cross-entropy, respectively.
We can see that
(1) \textbf{High judgment accuracy}:
Our method reaches a high classification accuracy on evaluation without increasing parameters.
As shown in the first line in Table~\ref{tab:compare}, the judgment accuracy and reason accuracy of the test set are \(74.4\pm 0.5\) and \(74.8 \pm 3.1\). These two accuracies on the test-random set are \(69.4 \pm 0.4\) and \(75.3 \pm 4.6\).
We can see that our result even surpasses to pure-classification models \wrt regarding the judgment and reason as two independent classification problems and jointly train via multitasking.
Meanwhile, under the setting of cross-entropy loss,
the judgment accuracy is larger than \emph{Reason-NoReg} which has no regularization.
This demonstrates the effectiveness of the gradient penalty.
(2) \textbf{High reason accuracy}: For methods like IFIV and Reason-NoReg,
they can bring us some explanations or feedback.
However, their results are less likely to be fully aligned to expert thoughts.
(3) \textbf{Good generalization}: From the good performance on test-random sets,
we see that our result still achieves the highest accuracy among the compared methods, which means our approach enjoys good generalizability.

As discussed in the Section~\ref{sec:related},
the embedding methods are mainly adopted in previous works.
We conducted a study to show that embedding may fail to distinguish \good, \normal, and \bad,
which makes it not suitable for our task.
There are two basic ways to represent evaluation:
absolute rating and relative embedding.
Embedding methods assume that there is a common compatibility space for fashion items.
The idea is to pull the representations of positive pairs closer and push the negative pairs farther (positive pairs are those that appear in dataset e.g. Polyvore~\cite{han2017learning} and FashionVC~\cite{song2017neurostylist}, and negative pairs are random combinations).
%    The annotated data for this method is cheap to get, however we discovered the embedding results fail to distinguish good, normal and bad.
We built an embedding model consisting of two ResNet18 networks which respectively embeds top and bottom to the compatibility space. The compatibility is measured by the cosine distance, and the model is trained using the BPR loss.
After training, the average distance between top and bottom of bad is 0.464, far less than that of normal (0.562) and good (0.574).
This result is against with the assumption that more compatible items have a smaller distance.
% which shows the necessity of creating evaluation datasets with grading annotations.

\paragraph{Qualitative analysis.}
We show some examples with more details in Figure~\ref{fig:results}.
Take the first mini-table as an example.
The ground-truth and predicted judgment for this outfit are both \good while the ground-truth reason is color.
The scores computed by Equation~\eqref{eq:f} (the second row with label \emph{G}) when set \(j^{gt}\) = \good are 0.647 in color, 0.098 in print,
and 0.045 in attribute, which means color is the predicted reason.
Similarly, the scores calculated by Equation~\eqref{eq:contrib} (the first row with label \emph{C}) are shown.
The red number means the predicted reason is wrong while green one means correct.
The first two rows are examples of \good, and \bad, respectively.
We can see that our model provides more accurate judgment with the operation of taking \emph{difference}.
Further, to demonstrate the model indeed learned the concept of color and print,
we change the color or print of a good outfit in the last row.
The prediction changes accordingly. Furthermore,
the reason reflects the changing factor.
This demonstrates our model indeed learned the reason for its judgment.

Additionally, we present the final performance of our outfit evaluation system in Figure~\ref{fig:visual}.
Given an outfit, an absolute judgment with convincing explanation can be provided to users.
Take the outfit at the fifth line in Figure~\ref{fig:visual} for example.
This outfit is \bad,
because the pink yarrow color in top and cadmium green color in the bottom is wrong color matching.
Meanwhile, our method could also be applied in outfit recommendation scene by simply scaling the logit or the probability of the judgment as of the relative score for recommendation sort.
This kind of explanation framework took the suggestions of fashion experts as reference.
%  and thus we believe it would be useful in practical applications.

\begin{figure}[t]
    \begin{center}
        \includegraphics[width=1\linewidth]{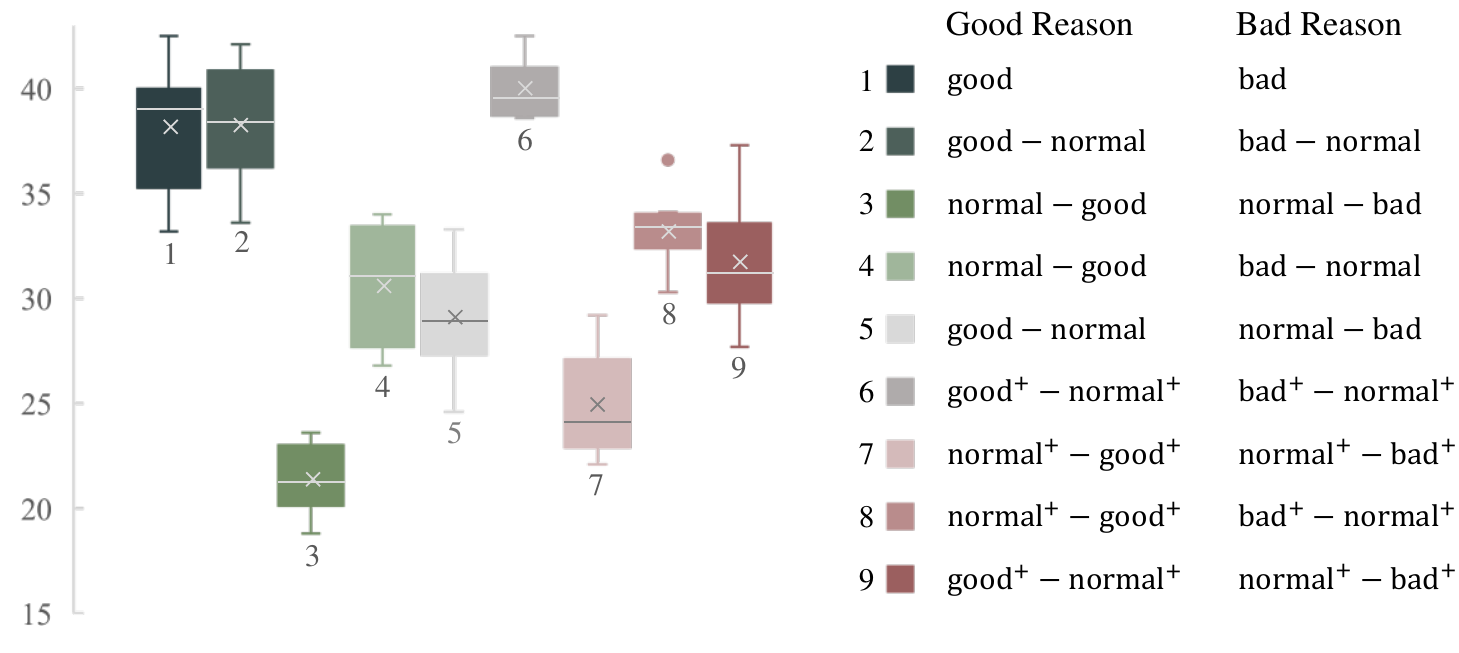}
    \end{center}
    \caption{\small
        The reason validation accuracy for different formulations in the unsupervised setting,
        \ie \emph{without} any reason-related regularizations.
        All experiments are repeated \(10\) times.
        Here the notation ``\(\text{good}\)'' represents \(\contrib_{good}\) and ``\(\text{good}^+\)'' is shorthand for \(\contrib_{good}^+\), \etc.
        We can see that the 6-th formulation \emph{positive contribution difference} has the greatest accuracy on average.
    }\label{fig:observation}
\end{figure}

\begin{figure}[t]
    \begin{center}
        \includegraphics[width=1\linewidth]{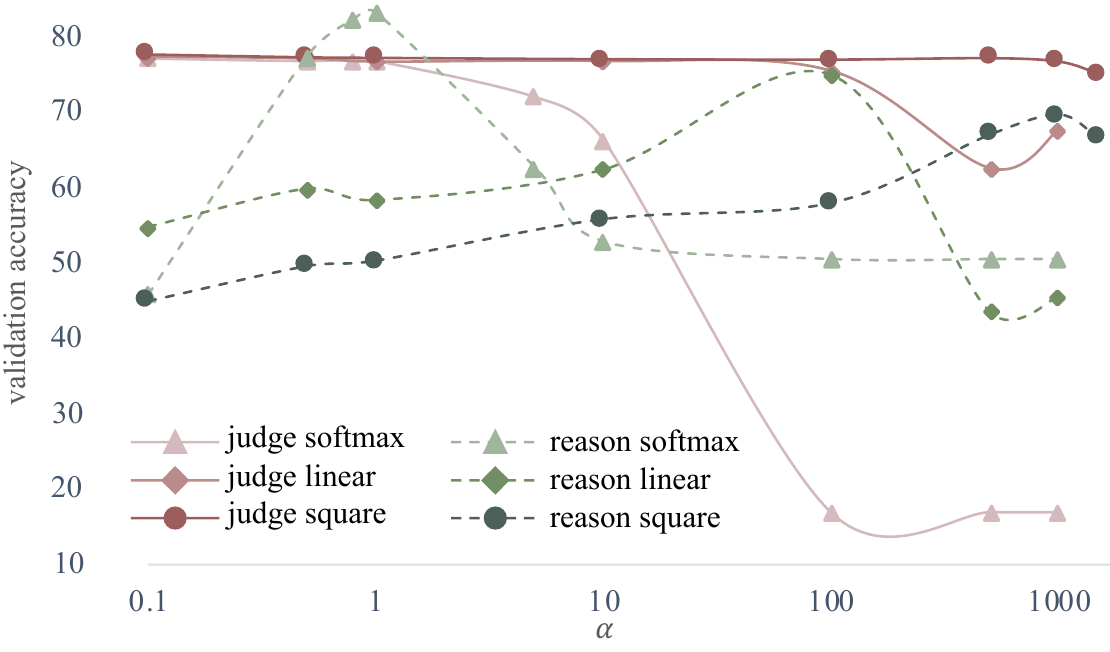}
    \end{center}
    \caption{\small
        The change of judgment accuracies and reason accuracies with different regularization coefficient \(\alpha\).
        We show three regularization methods: cross-entropy, linear and square.
        All experiments were repeated \(5\) times,
        and we draw the average validation accuracies on the figure.
        % All of them demonstrates the trade-off between the judgment accuracy and the reason accuracy.
    }\label{fig:alpha}
\end{figure}

\subsection{Ablation study}
\paragraph{Contribution formulations.}
We explore other candidate formulations for the \emph{reasons} in fashion compatibility evaluation.
The network in Figure~\ref{fig:pipeline} was trained \emph{without} any reason regularizations,
\ie using an unsupervised treatment for reasons.
After the training,
we analyze the reason accuracy for different reason formulations in Figure~\ref{fig:observation}.
The first one is to directly use the contribution defined in Equation~\eqref{eq:contrib} and we use it as the reference performance.
Especially, Grad-CAM is equivalent to the first one when we take the maximum for the major reason.
We see that the contribution difference (the 2nd formulation) and the positive contribution difference (the 6th formulation) are slightly better than the baseline.
Particularly, the positive contribution difference has lower variance and thus we adopt the 6th one as the formulation for reason.

\paragraph{Regularizers.}
% Meanwhile, we study the effect of the regularization under different conditions.
Meanwhile, we conduct experiments on three different forms of regularization:
(a) cross-entropy, (b) linear, and (c) square.
The judgment accuracies and the reason accuracies of these formulations are plotted in Figure~\ref{fig:alpha},
in which the regularization coefficient varies.
For all three regularization methods,
the trade-off between judgment accuracy and the reason accuracy.
When \(\alpha\) increases,
the reason accuracy generally improves.
The judgment accuracy can maintain at the same level for some time,
and finally drops when \(\alpha\) becomes too large.
The cross-entropy regularization is most sensitive to the change of hyper-parameter,
and reaches the best judgment \& reason accuracy trade-off at \(\alpha=1\).
The best \(\alpha\) value for linear and square formulations are \(100\) and \(1000\), respectively.

%%%%%%%%%%%%%%%%%%%%%%%%%%%%%%%%%%%%%%%%%%%%%%%%%%%%%%%%%%%%%%%%%
% CONCLUSION
%%%%%%%%%%%%%%%%%%%%%%%%%%%%%%%%%%%%%%%%%%%%%%%%%%%%%%%%%%%%%%%%%
\section{Conclusion}\label{sec:conclusion}
In this work, we present an outfit evaluation system with the feedback consisting of an absolute judgment together with an explanation.
In particular, the interpretation of neural networks produced by our method is aligned with expert annotated reasons by adding a gradient penalty regularization.
We use the difference of contributions to model the interaction of the two judgments of \good and \bad with the \normal judgment.
With the reason regularization term imposed on the loss,
we can make the network more interpretable without losing its performance.
Comprehensive experiments conducted on the newly annotated \evaluationthree show the effectiveness of the proposed method.

{\small
\bibliographystyle{ieee_fullname}
\bibliography{egbib}

\begin{thebibliography}{10}\itemsep=-1pt

\bibitem{al2017fashion}
Ziad Al-Halah, Rainer Stiefelhagen, and Kristen Grauman.
\newblock Fashion forward: Forecasting visual style in fashion.
\newblock In {\em Proceedings of the IEEE International Conference on Computer
  Vision}, pages 388--397, 2017.

\bibitem{al2017contextual}
Maruan Al-Shedivat, Avinava Dubey, and Eric~P Xing.
\newblock Contextual explanation networks.
\newblock {\em arXiv preprint arXiv:1705.10301}, 2017.

\bibitem{chattopadhay2018grad}
Aditya Chattopadhay, Anirban Sarkar, Prantik Howlader, and Vineeth~N
  Balasubramanian.
\newblock Grad-cam++: Generalized gradient-based visual explanations for deep
  convolutional networks.
\newblock In {\em 2018 IEEE Winter Conference on Applications of Computer
  Vision}, pages 839--847. IEEE, 2018.

\bibitem{chen2018dress}
Long Chen and Yuhang He.
\newblock Dress fashionably: Learn fashion collocation with deep mixed-category
  metric learning.
\newblock In {\em Thirty-Second AAAI Conference on Artificial Intelligence},
  2018.

\bibitem{chen2018visually}
Xu Chen, Yongfeng Zhang, Hongteng Xu, Yixin Cao, Zheng Qin, and Hongyuan Zha.
\newblock Visually explainable recommendation.
\newblock {\em preprint arXiv:1801.10288}, 2018.

\bibitem{Cucurull_2019_CVPR}
Guillem Cucurull, Perouz Taslakian, and David Vazquez.
\newblock Context-aware visual compatibility prediction.
\newblock In {\em The IEEE Conference on Computer Vision and Pattern
  Recognition (CVPR)}, June 2019.

\bibitem{cui2019dressing}
Zeyu Cui, Zekun Li, Shu Wu, Xiaoyu Zhang, and Liang Wang.
\newblock Dressing as a whole: Outfit compatibility learning based on node-wise
  graph neural networks.
\newblock {\em arXiv preprint arXiv:1902.08009}, 2019.

\bibitem{eckman1995aesthetic}
Molly Eckman and Janet Wagner.
\newblock Aesthetic aspects of the consumption of fashion design: The
  conceptual and empirical challenge.
\newblock {\em ACR North American Advances}, 1995.

\bibitem{feng2018interpretable}
Zunlei Feng, Zhenyun Yu, Yezhou Yang, Yongcheng Jing, Junxiao Jiang, and Mingli
  Song.
\newblock Interpretable partitioned embedding for customized multi-item fashion
  outfit composition.
\newblock In {\em Proceedings of the 2018 ACM on International Conference on
  Multimedia Retrieval}, pages 143--151. ACM, 2018.

\bibitem{echolook}
Samuel Gibbs.
\newblock Amazon unveils echo look, a selfie camera to help you choose what to
  wear.
\newblock
  \url{https://www.theguardian.com/technology/2017/apr/26/amazon-echo-look-webcam-choose-fashion-outfits-alexa-smart-selfie-camera},
  2017.

\bibitem{han2017learning}
Xintong Han, Zuxuan Wu, Yu-Gang Jiang, and Larry~S Davis.
\newblock Learning fashion compatibility with bidirectional lstms.
\newblock In {\em Proceedings of the 2017 ACM on Multimedia Conference}, pages
  1078--1086. ACM, 2017.

\bibitem{he2016deep}
Kaiming He, Xiangyu Zhang, Shaoqing Ren, and Jian Sun.
\newblock Deep residual learning for image recognition.
\newblock In {\em Proceedings of the IEEE conference on computer vision and
  pattern recognition}, pages 770--778, 2016.

\bibitem{hou2019explainable}
Min Hou, Le Wu, Enhong Chen, Zhi Li, Vincent~W Zheng, and Qi Liu.
\newblock Explainable fashion recommendation: A semantic attribute region
  guided approach.
\newblock {\em Twenty-Eight International Joint Conference on Artificial
  Intelligence}, 2019.

\bibitem{hsiao2017learning}
Wei-Lin Hsiao and Kristen Grauman.
\newblock Learning the latent “look”: Unsupervised discovery of a
  style-coherent embedding from fashion images.
\newblock In {\em 2017 IEEE International Conference on Computer Vision
  (ICCV)}, pages 4213--4222. IEEE, 2017.

\bibitem{hsiao2018creating}
Wei-Lin Hsiao and Kristen Grauman.
\newblock Creating capsule wardrobes from fashion images.
\newblock In {\em Proceedings of the IEEE Conference on Computer Vision and
  Pattern Recognition}, pages 7161--7170, 2018.

\bibitem{kang2018complete}
Wang-Cheng Kang, Eric Kim, Jure Leskovec, Charles Rosenberg, and Julian
  McAuley.
\newblock Complete the look: Scene-based complementary product recommendation.
\newblock {\em arXiv preprint arXiv:1812.01748}, 2018.

\bibitem{kim2018introduction}
Been Kim and F Doshi-Velez.
\newblock Introduction to interpretable machine learning.
\newblock {\em Proceedings of the CVPR Tutorial on Interpretable Machine
  Learning for Computer Vision. https://interpretablevision. github.
  io/index\_cvpr2018. html}, 2018.

\bibitem{li2017mining}
Yuncheng Li, Liangliang Cao, Jiang Zhu, and Jiebo Luo.
\newblock Mining fashion outfit composition using an end-to-end deep learning
  approach on set data.
\newblock {\em IEEE Transactions on Multimedia}, 19(8):1946--1955, 2017.

\bibitem{lin2018explainable}
Yujie Lin, Pengjie Ren, Zhumin Chen, Zhaochun Ren, Jun Ma, and Maarten de
  Rijke.
\newblock Explainable fashion recommendation with joint outfit matching and
  comment generation.
\newblock {\em arXiv preprint arXiv:1806.08977}, 2018.

\bibitem{mcauley2015image}
Julian McAuley, Christopher Targett, Qinfeng Shi, and Anton Van Den~Hengel.
\newblock Image-based recommendations on styles and substitutes.
\newblock In {\em Proceedings of the 38th International ACM SIGIR Conference on
  Research and Development in Information Retrieval}, pages 43--52. ACM, 2015.

\bibitem{melis2018towards}
David~Alvarez Melis and Tommi Jaakkola.
\newblock Towards robust interpretability with self-explaining neural networks.
\newblock In {\em Advances in Neural Information Processing Systems}, pages
  7786--7795, 2018.

\bibitem{montavon2018methods}
Gr{\'e}goire Montavon, Wojciech Samek, and Klaus-Robert M{\"u}ller.
\newblock Methods for interpreting and understanding deep neural networks.
\newblock {\em Digital Signal Processing}, 73:1--15, 2018.

\bibitem{oramas2016modeling}
Jose Oramas and Tinne Tuytelaars.
\newblock Modeling visual compatibility through hierarchical mid-level
  elements.
\newblock {\em arXiv preprint arXiv:1604.00036}, 2016.

\bibitem{pantone}
Pantone.
\newblock Pantone color, chips \& color guides | color inspiration.
\newblock \url{https://www.pantone.com/}, 2019.

\bibitem{plumb2019regularizing}
Gregory Plumb, Maruan Al-Shedivat, Eric Xing, and Ameet Talwalkar.
\newblock Regularizing black-box models for improved interpretability.
\newblock {\em arXiv preprint arXiv:1902.06787}, 2019.

\bibitem{selvaraju2017grad}
Ramprasaath~R Selvaraju, Michael Cogswell, Abhishek Das, Ramakrishna Vedantam,
  Devi Parikh, and Dhruv Batra.
\newblock Grad-cam: Visual explanations from deep networks via gradient-based
  localization.
\newblock In {\em Proceedings of the IEEE International Conference on Computer
  Vision}, pages 618--626, 2017.

\bibitem{shih2018compatibility}
Yong-Siang Shih, Kai-Yueh Chang, Hsuan-Tien Lin, and Min Sun.
\newblock Compatibility family learning for item recommendation and generation.
\newblock In {\em Thirty-Second AAAI Conference on Artificial Intelligence},
  2018.

\bibitem{simo2015neuroaesthetics}
Edgar Simo-Serra, Sanja Fidler, Francesc Moreno-Noguer, and Raquel Urtasun.
\newblock Neuroaesthetics in fashion: Modeling the perception of
  fashionability.
\newblock In {\em Proceedings of the IEEE Conference on Computer Vision and
  Pattern Recognition}, pages 869--877, 2015.

\bibitem{simonyan2013deep}
Karen Simonyan, Andrea Vedaldi, and Andrew Zisserman.
\newblock Deep inside convolutional networks: Visualising image classification
  models and saliency maps.
\newblock {\em arXiv preprint arXiv:1312.6034}, 2013.

\bibitem{song2017neurostylist}
Xuemeng Song, Fuli Feng, Jinhuan Liu, Zekun Li, Liqiang Nie, and Jun Ma.
\newblock Neurostylist: Neural compatibility modeling for clothing matching.
\newblock In {\em Proceedings of the 2017 ACM on Multimedia Conference}, pages
  753--761. ACM, 2017.

\bibitem{springenberg2014striving}
Jost~Tobias Springenberg, Alexey Dosovitskiy, Thomas Brox, and Martin
  Riedmiller.
\newblock Striving for simplicity: The all convolutional net.
\newblock {\em arXiv preprint arXiv:1412.6806}, 2014.

\bibitem{tangseng2019toward}
Pongsate Tangseng and Takayuki Okatani.
\newblock Toward explainable fashion recommendation.
\newblock {\em arXiv preprint arXiv:1901.04870}, 2019.

\bibitem{vasileva2018learning}
Mariya~I Vasileva, Bryan~A Plummer, Krishna Dusad, Shreya Rajpal, Ranjitha
  Kumar, and David Forsyth.
\newblock Learning type-aware embeddings for fashion compatibility.
\newblock In {\em Proceedings of the European Conference on Computer Vision
  (ECCV)}, pages 390--405, 2018.

\bibitem{veit2015learning}
Andreas Veit, Balazs Kovacs, Sean Bell, Julian McAuley, Kavita Bala, and Serge
  Belongie.
\newblock Learning visual clothing style with heterogeneous dyadic
  co-occurrences.
\newblock In {\em Proceedings of the IEEE International Conference on Computer
  Vision}, pages 4642--4650, 2015.

\bibitem{wang2019outfit}
Xin Wang, Bo Wu, Yun Ye, and Yueqi Zhong.
\newblock Outfit compatibility prediction and diagnosis with multi-layered
  comparison network.
\newblock In {\em Proceedings of the 2019 ACM on Multimedia Conference}. ACM,
  2019.

\bibitem{yang2019interpretable}
Xun Yang, Xiangnan He, Xiang Wang, Yunshan Ma, Fuli Feng, Meng Wang, and
  Tat-Seng Chua.
\newblock Interpretable fashion matching with rich attributes.
\newblock SIGIR, 2019.

\bibitem{zhou2016learning}
Bolei Zhou, Aditya Khosla, Agata Lapedriza, Aude Oliva, and Antonio Torralba.
\newblock Learning deep features for discriminative localization.
\newblock In {\em Proceedings of the IEEE conference on computer vision and
  pattern recognition}, pages 2921--2929, 2016.

\bibitem{zou2019foco}
Xingxing Zou, Wai~Keung Wong, Can Gao, and Jie Zhou.
\newblock Foco system: a tool to bridge the domain gap between fashion and
  artificial intelligence.
\newblock {\em International Journal of Clothing Science and Technology}, 2019.

\end{thebibliography}
}

\end{document}